\documentclass[]{bytedance_seed}
\usepackage[T1]{fontenc}
\usepackage[toc,page,header]{appendix}

\usepackage{wrapfig}
\usepackage{multirow}
\usepackage{makecell}
\usepackage{xcolor}
\usepackage{colortbl}

\usepackage{amsmath,amsfonts,bm}

\usepackage{xspace}

\newcommand{\markedblanklines}[2][\textbullet]{
  \par
  \begingroup
  \setlength{\parskip}{0pt}
  \setlength{\parindent}{0pt}
  \count0=0
  \loop
    \ifnum\count0<#2
      #1\hspace{1em}\rule{0pt}{\baselineskip}\par
      \advance\count0 by 1
  \repeat
  \endgroup
}

\def\eqref#1{equation~\ref{#1}}

\def\1{\bm{1}}

\DeclareMathAlphabet{\mathsfit}{\encodingdefault}{\sfdefault}{m}{sl}
\SetMathAlphabet{\mathsfit}{bold}{\encodingdefault}{\sfdefault}{bx}{n}

\definecolor{colorfirst}{rgb}{.866,.945, 0.831}
\definecolor{colorsecond}{rgb}{1, 0.98, 0.83}
\definecolor{colorthird}{rgb}{0.76, 0.87, 0.92}
\definecolor{colorcite}{rgb}{0.212, 0.490, 0.741}

\usepackage{cleveref}
\crefname{figure}{Fig.}{Figs.}
\crefname{table}{Tab.}{Tabs.}
\crefname{equation}{Eq.}{Eqs.}
\crefname{section}{Sec.}{Secs.}

\title{Robust 4D Visual Geometry Transformer with Uncertainty-Aware Priors }

\author[3]{Ying Zang}
\author[3]{Yidong Han}
\author[1]{Chaotao Ding}
\author[1]{Yuanqi Hu}
\author[4]{Deyi Ji}
\author[4]{Qi Zhu}
\author[5]{Xuanfu Li} 
\author[5]{Jin Ma}
\author[2]{Lingyun Sun}
\author[1,2,\dagger]{Tianrun Chen}
\author[6]{Lanyun Zhu}

\affiliation[1]{KOKONI 3D, Moxin Technology}
\affiliation[2]{Zhejiang University}
\affiliation[3]{Huzhou University}
\affiliation[4]{Univeristy of Science and Technology
of China}
\affiliation[5]{Huawei}
\affiliation[6]{Tongji University}

\contribution[\dagger]{Project Lead}

\abstract{

Reconstructing dynamic 4D scenes is an important yet challenging task. While 3D foundation models like VGGT excel in static settings, they often struggle with dynamic sequences where motion causes significant geometric ambiguity. To address this, we present a framework designed to disentangle dynamic and static components by modeling uncertainty across different stages of the reconstruction process. Our approach introduces three synergistic mechanisms:   (1) Entropy-Guided Subspace Projection, which leverages information-theoretic weighting to adaptively aggregate multi-head attention distributions, effectively isolating dynamic motion cues from semantic noise; (2) Local-Consistency Driven Geometry Purification, which enforces spatial continuity via radius-based neighborhood constraints to eliminate structural outliers; and (3) Uncertainty-Aware Cross-View Consistency, which formulates multi-view projection refinement as a heteroscedastic maximum likelihood estimation problem, utilizing depth confidence as a probabilistic weight.  Experiments on dynamic benchmarks show that our approach outperforms current state-of-the-art methods, reducing Mean Accuracy error by 13.43\% and improving segmentation F-measure by 10.49\%. Our framework maintains the efficiency of feed-forward inference and requires no task-specific fine-tuning or per-scene optimization.

}

\correspondence{Tianrun Chen, Lanyun Zhu}

\begin{document}

\maketitle

\footnotetext[1]{We thank Jianyuan Wang for his insightful discussions. We acknowledge the support from Hisilicon, the ZJU Kunpeng \& Ascend Center of Excellence, and the Dream Set Off - Kunpeng \& Ascend Seed Program.}

\section{Introduction}

The reconstruction of 4D scenes, jointly recovering the 3D geometry, camera motion, and object dynamics from multi-view video, is a central problem in computer vision with broad applications in autonomous driving \cite{kitti}, robotics \cite{dynamicfusion}, and immersive media \cite{kerbl20233d}. Recent 3D foundation models, particularly the Visual Geometry Grounded Transformer (VGGT) \cite{vggt}, have achieved remarkable success in static multi-view reconstruction by leveraging global attention mechanisms to aggregate cross-view geometric evidence. However, the real world is inherently dynamic, and the deployment of these models in dynamic environments reveals a fundamental limitation: the inability to robustly decouple moving objects from the static background.

This limitation stems from a deep structural contradiction in multi-view geometry. Traditional pipelines rely heavily on the assumption of multi-view rigidity and photometric constancy. When dynamic objects violate these assumptions, they introduce severe geometric residuals. If these residuals are treated uniformly with static correspondences, they inevitably corrupt the reconstruction process, leading to catastrophic pose drift and severe artifacts in the reconstructed 3D scene representations. Therefore, explicitly masking out dynamic regions is a fundamental requirement for unbiased pose estimation and robust geometry reconstruction.

While existing methods have made progress in 4D reconstruction, they often fall short in practical scenarios. Optimization-based approaches \cite{kopf2021robust} achieve high-quality results through iterative refinement but incur substantial computational overhead, making them unsuitable for long sequences. Learning-based methods attempt to learn dynamic-aware representations but require fine-tuning on large-scale dynamic datasets, limiting their generalization to out-of-distribution scenes. Training-free variants, such as Easi3R \cite{easi3r}, represent a promising direction by extracting dynamic cues directly from pre-trained models. However, they are often confined to pairwise architectures and produce coarse masks lacking temporal consistency and sharp boundaries.

We observe that the fundamental issue underlying these
limitations is the absence of principled uncertainty quantification across the reconstruction pipeline. The dynamic-static decoupling problem is inherently multi-scale and
multi-space: it involves feature-level information mixing,
geometry-level structural corruption, and constraint-level
projection uncertainty. Any method that addresses only one
of these error sources leaves the others uncontrolled, leading to suboptimal results.

To address these interconnected challenges, we propose
a unified hierarchical framework that systematically models
uncertainty at three complementary levels. First, at the feature representation level, we introduce an Entropy-Guided
Subspace Projection mechanism. Motivated by the observation that different attention heads capture information with vastly different reliability, we employ an entropy driven mechanism to adaptively weight heads based on their
information content, amplifying genuine motion cues while
suppressing diffuse noise. Second, at the geometric level,
we propose a Local-Consistency Driven Geometry Purification mechanism. This enforces local spatial continuity by removing isolated outlier points that lack sufficient neighborhood support, thereby purifying the dynamic
point cloud. Finally, at the constraint level, we introduce an
Uncertainty-Aware Cross-View Consistency mechanism.
By formulating the projection loss as a heteroscedastic maximum likelihood estimation problem, we incorporate depth
confidence as a probabilistic weight, effectively mitigating
the impact of unreliable observations in occluded or texture-less regions. Extensive experiments across multiple dynamic benchmarks demonstrate that our approach achieves state-of-the-art performance, significantly outperforming existing methods in both dynamic object segmentation and dense geometric reconstruction.

\section{Related Work}

\subsection{3D Foundation Models}

Modern visual perception has witnessed a profound shift toward scalable, data-driven representation learning and unified modeling paradigms\cite{arnold2022map,zhu2024llafs,huang2018deepmvs,zhu2025skysense,zhu2024ibd,sstkd_pami,dlpl,zhu2025not,fastvggt,sstkd,reizenstein2021common,pptformer}, yielding substantial advances across a broad spectrum of computer vision tasks ranging from semantic understanding to geometric inference. The paradigm of 3D reconstruction has been revolutionized by feed-forward foundation models \cite{li2018megadepth,zhu2025replay,wang2025pi,zhu2025llafs++,cagcn,zhu2025cpcf,ji2026view,vggt}. DUSt3R \cite{dust3r} pioneered large-scale pretraining
for pose-free dense 3D reconstruction from unposed image pairs, treating geometry estimation as a direct regression problem. MASt3R \cite{MASt3R} further enhanced correspondence quality by introducing matching-aware representations. Building upon these pairwise methods, VGGT \cite{vggt} extended the paradigm to multi-view inputs via a unified global attention mechanism, enabling joint prediction of camera poses and point maps. Recent variants have explored streaming architectures \cite{stream3r}, token-merging acceleration \cite{fastvggt}, and permutation-equivariant designs \cite{wang2025pi}. However, a fundamental limitation persists: these models are explicitly trained under the assumption of static environments.
When deployed in the real world, moving objects violate the
underlying epipolar constraints, leading to severe attention
pollution and catastrophic pose drift.

\subsection{4D Scene Reconstruction} 

Reconstructing dynamic scenes from monocular video is a long-standing challenge.
Early methods relied heavily on RGB-D sensors  or strong category-specific priors \cite{dynamicfusion,urur,volumedeform,zhu2021learning,schonberger2016pixelwise,zhu2025popen,zhu2025retrv,gpwformer,xiao2025spatialtrackerv2}. Recent optimization-based approaches, such as Robust CVD \cite{kopf2021robust} 
and MegaSAM \cite{megasam}, achieve high-quality results through
iterative test-time optimization but incur substantial computational overhead, making them unsuitable for long sequences or real-time applications. To address efficiency,
learning-based methods like MonST3R \cite{monst3r}, DAS3R \cite{das3r},
and CUT3R \cite{cut3r} attempt to learn dynamic-aware representations directly. However, they require extensive finetuning on large-scale dynamic datasets, which limits their
generalization to out-of-distribution scenes. Training-free
variants, such as Easi3R \cite{easi3r}, represent a promising direction by extracting dynamic cues directly from pretrained
models. Yet, they are often limited to pairwise architectures and produce coarse masks. Concurrently, PAGE-4D \cite{zhou2025page} addresses this issue through architectural disentanglement, introducing an explicit mask prediction module and a dynamics-aware aggregator that requires targeted fine-tuning of specific attention layers to physically separate dynamic and static features. In contrast, our framework
approaches the problem through uncertainty-driven disentanglement. We posit that dynamic regions fundamentally
manifest as high-uncertainty areas in multi-view geometry
(e.g., diffuse attention variance, local geometric inconsistency, and low projection confidence). By explicitly modeling and propagating this uncertainty across the feature, geometry, and constraint levels, our framework naturally isolates dynamic objects without requiring architectural modifications or domain-specific retraining.

\subsection{Uncertainty in Multi-View Geometry}

Modeling uncertainty is crucial for robust perception \cite{su2022uncertainty,liao2024multi,zhang2023geomvsnet,chen2025learning, fervers2023uncertainty}. Bayesian
approaches to depth estimation \cite{kendall2017uncertainties} distinguish between aleatoric (data-dependent) and epistemic (model dependent) uncertainty. In multi-view geometry, heteroscedastic regression \cite{nix1994estimating} has been used to weight observations by their inverse variance, mitigating the impact of noisy measurements. While these principles have been applied to static depth estimation, their application to dynamic scene reconstruction remains underexplored. Our work bridges this gap by extending heteroscedastic modeling to the dynamic masking problem, providing a unified probabilistic framework that spans feature aggregation, geometric purification, and cross-view consistency

\begin{figure*}[t]
  \centering
  \includegraphics[width=\linewidth]{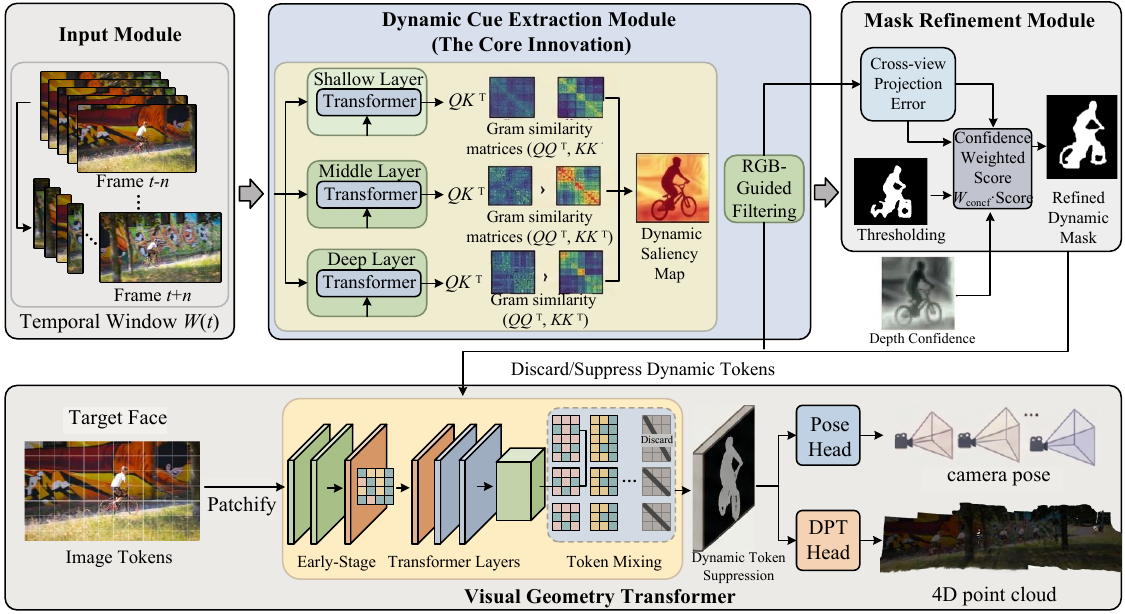} 
  \caption{Overview of the proposed framwwork. }
  \label{fig:arch}
\end{figure*}

\section{Method}

In this section, we detail our hierarchical framework for
dynamic-static decoupling. We first analyze the impact of
dynamic objects on the VGGT architecture, and
then introduce our three core mechanisms: Entropy-Guided
Subspace Projection, Local-Consistency Driven
Geometry Purification, and Uncertainty-Aware
Cross-View Consistency.

\subsection{Preliminaries}

\subsubsection{Visual Geometry Grounded Transformer}

The core of our framework is built upon the VGGT (Visual Geometry Grounded Transformer) \cite{vggt} architecture, which treats multi-view 3D reconstruction as a unified sequence-to-sequence prediction task. Unlike traditional methods that regress dense depth maps directly, VGGT represents the underlying 4D scene through a set of geometric primitives.

Specifically, given an input video sequence $\mathcal{I} = \{I_1, \dots, I_T\}$, the model tokenizes each frame into patches and processes them through a global attention-based Transformer. The fundamental output of this process is a dense map of 3D primitives $\mathbf{P}$. Each primitive $p_{i,t} \in \mathbf{P}$ at pixel $i$ and time $t$ is parameterized by its 3D coordinates $\mathbf{X} \in \mathbb{R}^3$ and a local geometric descriptor. 

A key feature of the VGGT architecture is the use of a cross-view attention mechanism, which enables the model to implicitly learn geometric correlations between different frames. For any two views, the relative camera pose $(R, t)$ and the 3D geometry are jointly inferred by minimizing the discrepancy between the projected primitives and their corresponding observations.

Under static conditions, geometry estimation relies on implicitly modeling the correspondence
between a reference-frame homogeneous pixel $\mathbf{x}_r$ and its
target-frame counterpart $\mathbf{x}_t$. This relationship is governed
by the standard rigid-scene projection equation:

\begin{equation}
    \mathbf{x}_t = \mathbf{K}[\mathbf{R}_{t \leftarrow r} D_r(\mathbf{x}_r) \mathbf{K}^{-1} \mathbf{x}_r + \mathbf{t}_{t \leftarrow r}]
\end{equation}

where $\mathbf{K}$ is the intrinsic matrix, $D_r$ is the depth, and $(\mathbf{R}_{t \leftarrow r}, t_{t \leftarrow r})$ represents the relative pose. This
static assumption allows models to fit an essential matrix
$\mathbf{E} = [\mathbf{t}_{t \leftarrow r}]\mathbf{R}_{t \leftarrow r}$ that enforces the epipolar constraint $\tilde{\mathbf{x}}^{\top}_{t} \mathbf{E}\tilde{\mathbf{x}}_r = 0$. Our proposed method leverages this primitive-based representation but introduces hierarchical uncertainty modeling to overcome its inherent limitations in dynamic environments.

\subsubsection{The Necessity of Dynamic Masking}
To reveal why explicit dynamic masking is indispensable for foundation models such as VGGT, we follow the analytical frameworks of VGGT4D \cite{vggt4d} and Page-4D \cite{zhou2025page} to examine the underlying multi-view geometric constraints.

However, in dynamic scenes, object motion violates this assumption. The projection equation must be modified to account for the dynamic displacement $\mathbf{M}_{t \leftarrow r}$ induced by the moving object:

\begin{equation}
    \mathbf{x}_t = \mathbf{K}[\mathbf{R}_{t \leftarrow r} D_r(x_r) \mathbf{K}^{-1} \mathbf{x}_r + \mathbf{t}_{t \leftarrow r}] + \mathbf{K} \mathbf{M}_{t \leftarrow r}
\end{equation}

This dynamic displacement introduces a non-vanishing residual into the epipolar constraint:
\begin{equation}
    \delta(x_r) \equiv \tilde{\mathbf{x}}_t^\top \mathbf{E} \tilde{\mathbf{x}}_r \approx \frac{1}{Z_r} \mathbf{n}(x_r)^\top \Delta \mathbf{X}_{\perp}(\mathbf{x}_r)
\end{equation}
where $\mathbf{n}(\mathbf{x}_r)$ is the unit normal of the epipolar line and $\Delta \mathbf{X}_{\perp}(\mathbf{x}_r)$ is the component of the dynamic displacement perpendicular to that line.

In the VGGT architecture, cross-view geometry is inferred through a global attention mechanism. The presence
of the residual $\delta(\mathbf{x}_r)$ means that dynamic features act as severe distractors. The softmax operation forces the model to
distribute attention mass, and the moving objects cause the
attention distribution to become diffuse or incorrectly localized. Because camera pose estimation is highly brittle to dynamic motion—where even small residuals can corrupt the
essential matrix fitting—explicitly identifying and masking
out these dynamic regions is not merely a post-processing
step, but a fundamental mathematical requirement to preserve the integrity of the global attention mechanism and
ensure unbiased pose recovery

\subsection{Entropy-Guided Subspace Projection}

 In the VGGT architecture, the global attention mechanism computes a set of attention matrices ${\mathbf{A}^{(h)}}^H_{h=1}$ across $H$ heads. Standard approaches aggregate these via simple averaging, $\bar{\mathbf{A}} = \frac{1}{H} \sum_h \mathbf{A}(h)$. However, this implicitly assumes that all heads contribute equally to
the dynamic region detection. In reality, different heads
capture information with vastly different reliability—some
heads focus sharply on genuine motion cues, while others
produce diffuse, noisy, or redundant signals.

 We formulate the attention aggregation
from an information-theoretic perspective. Let $\mathbf{A}^{(h)}(\mathbf{u})$ be
the spatial response map of the h-th attention head. We
measure the information content of each head using its spatial variance:

\begin{equation}
    V(\mathbf{A}^{(h)}) = \frac{1}{|\Omega|} \sum_{\mathbf{u} \in \Omega} (\mathbf{A}^{(h)}(\mathbf{u}) - \bar{\mathbf{A}}^{(h)})^2
\end{equation}

where $Ω$ is the spatial domain and $\bar{\mathbf{A}}^{(h)}$ is the mean response. Heads with high variance concentrate their attention on specific regions (indicating high discriminative
power and strong motion cues), while heads with low variance distribute attention uniformly (indicating diffuse, low-information noise).

To aggregate these heads, we propose a variance-driven
weighting scheme. We define the normalized weights directly proportional to their spatial variance:

\begin{equation}
    w_h = \frac{V(\mathbf{A}^{(h)})}{\sum_{k=1}^H V(\mathbf{A}^{(k)}) + \epsilon}
\end{equation}

where $\epsilon$ is a small constant for numerical stability. This formulation ensures that attention heads with low discriminative power are suppressed, while highly confident motion
cues are amplified, effectively projecting the multi-head attention onto an information-rich dynamic subspace.

\subsection{Local-Consistency Driven Geometry Purification}

 The initial saliency map is projected into
3D space to obtain a dynamic point cloud $\mathcal{P}_D = \{\mathbf{p}_i\}^M_{i=1}$. However, this point cloud inevitably contains structural outliers due to false positives in the attention maps, depth estimation errors, and occlusion artifacts. If left unaddressed, these isolated noise points will corrupt subsequent spatial
clustering and boundary refinement. We enforce local spatial consistency by requiring each valid dynamic point to
have sufficient neighborhood support.

 We enforce this regularity through a
radius-based local consistency constraint. For each point
$\mathbf{p}_i \in \mathcal{P}_D$, we define its local neighborhood $\mathcal{N}_r(\mathbf{p}_i)$ as the
set of points within a Euclidean radius $r$:

\begin{equation}
    \mathcal{N}_r(\mathbf{p}_i) = \{\mathbf{p}_j \in \mathcal{P}_D \setminus \{\mathbf{p}_i\} : \|\mathbf{p}_i - \mathbf{p}_j\|_2 \le r\}
\end{equation}

The local support degree $d_i$ is simply the cardinality of
this neighborhood: $d_i = |\mathcal{N}_r(\mathbf{p}_i)|$. We define a spatial filter
operator $\mathcal{F}_\tau$ that strictly enforces the local density lower
bound:

\begin{equation}
    \mathcal{F}_{\tau}(\mathbf{p}_i) = 
    \begin{cases} 
    \mathbf{p}_i & \text{if } d_i \ge \tau \\
    \emptyset & \text{if } d_i < \tau 
    \end{cases}
\end{equation}

where $\tau$ is the minimum support threshold. Crucially, to ensure scale-invariance across diverse environments, we implement this as an adaptive geometric filter. The radius $r$ is dynamically set to $2\%$ of the scene’s bounding box diagonal ($r = 0.02 · D_{scene}$), and the support threshold is set to $\tau = 16$. This operation effectively truncates the high frequency spatial noise, ensuring that the remaining points
adhere to the local continuity property of the underlying geometric structure. By purifying the point cloud, we provide
a stable foundation for the final cross-view refinement stage.

\subsection{Uncertainty-Aware Cross-View Consistency}

 The final stage refines the dynamic mask
by enforcing cross-view geometric consistency. For a 3D point $p$, its projection onto view $i$ yields a depth residual $r_{d,i} = d_{proj,i} -D_i(\mathbf{u}_i)$. Traditional methods compute the projection error and apply a uniform threshold. However, this assumes homoscedastic noise (constant variance),
which is fundamentally violated in dynamic scenes. Depth
uncertainty varies dramatically due to occlusions, textureless regions, and motion blur.

 Following Kendall and Gal \cite{kendall2017uncertainties}, we
formulate the cross-view consistency as a heteroscedastic Maximum Likelihood Estimation (MLE) problem. We model the depth observation as a Gaussian random variable:

\begin{equation}
    D_{i}(\mathbf{u})\sim\mathcal{N}(\mu_{i}(\mathbf{u}),\sigma_{i}^{2}(\mathbf{u}))
\end{equation}

where $µ_i$ is the true depth and $\sigma_i^2$ is the pixel-wise variance
representing the aleatoric uncertainty.

The joint negative log-likelihood for the multi-view projection of point $\mathbf{p}$ across visible views $V(\mathbf{p})$ is:

\begin{equation}
    \mathcal{L}_{MLE}(\mathbf{p}) = \sum_{i \in \mathcal{V}(p)} \left[ \frac{r_{d,i}^2}{2\sigma_i^2(\mathbf{u}_i)} + \frac{1}{2} \log \sigma_i^2(\mathbf{u}_i) \right]
\end{equation}

Minimizing this loss is equivalent to performing
precision-weighted regression. To implement this, we modify the prediction head of the VGGT architecture to output an additional channel representing the confidence logit $l_i(\mathbf{u})$. We map this logit to a strictly positive confidence value $C_i(\mathbf{u})$ using an exponential activation function:

\begin{equation}
    \mathbf{C}_i(\mathbf{u}) = 1 + \exp(l_i(\mathbf{u}))
\end{equation}

This activation ensures that the confidence is bounded below by 1, preventing numerical instability. We establish the mapping that this confidence is inversely proportional to the observation variance ($\mathbf{C}_i \propto \sigma_i^{-2}$). Thus, the refined dynamic score is computed as the confidence-weighted projection error:

\begin{equation}
    \mathcal{S}_{dyn}(\mathbf{p}) = \sum_{i \in \mathcal{V}(\mathbf{p})} \tilde{\mathbf{C}}_i(\mathbf{u}_i) \cdot (|d_{proj,i} - D_i(\mathbf{u}_i)| + \lambda |\mathbf{c}_{proj,i} - \mathbf{I}_i(\mathbf{u}_i)|)
\end{equation}

where $\tilde{\mathbf{C}}_i$ is the normalized confidence across all valid projections ($\sum \tilde{\mathbf{C}}_i = 1$), $\mathbf{c}_{proj,i}$ and $\mathbf{I}_i(\mathbf{u}_i)$ are the projected and sampled RGB colors, and $\lambda$ is a balancing weight (set to 1/3 in our implementation).

This formulation guarantees that highly uncertain regions (e.g., occlusions or textureless areas) naturally predict
lower confidence values, thereby receiving lower weights.
This prevents unreliable geometric observations from corrupting the dynamic mask refinement, acting as a robust
probabilistic filter on the projection process.

\section{Experiments}

\subsection{Experimental Setup}

\noindent
\textbf{Datasets.} Following the experimental setup of VGGT4D \cite{vggt4d}, we assess dynamic mask estimation on the DAVIS \cite{davis2016} dataset. For camera pose estimation, we evaluate our method on DyCheck \cite{dycheck}, specifically selecting several highly dynamic video sequences to test robustness. Our sampling strategy involves extracting frames every 4 steps for DyCheck to ensure consistent evaluation across different temporal scales. Furthermore, we utilize DyCheck to quantitatively evaluate point cloud reconstruction quality.

\textbf{Evaluation Metrics.} We provide a multi-dimensional assessment of our framework through three primary categories of metrics. (1) Reconstruction Quality: Evaluated on DyCheck, we report the Accuracy, Completeness, and Distance in terms of both Mean and Median values. These metrics quantify the precision of our uncertainty-aware geometry purification and projection refinement; (2) Pose Estimation: We measure the Absolute Trajectory Error (ATE) aggregated across all evaluation datasets. This metric is crucial for demonstrating that explicitly masking dynamic regions via entropy-guided cues prevents catastrophic pose drift in foundation models; (3) Segmentation Quality: For the DAVIS dataset, we report the Jaccard Mean (JM) and Boundary F-Measure (FM). These reflect the accuracy and boundary sharpness of the estimated dynamic masks, which are significantly enhanced by our entropy-guided subspace projection.

\noindent
\textbf{Baselines.} We evaluate our framework against a comprehensive suite of state-of-the-art pose-free 4D reconstruction methods, including MonST3R , DAS3R , CUT3R , Easi3R, and VGGT4D. Our comparisons utilize the VGGT4D baseline to benchmark improvements in reconstruction quality, pose estimation, and dynamic segmentation performance.

\begin{table*}[t]
\centering
\caption{Comparison with state-of-the-art methods on DyCheck dataset. Best results in bold, second best underlined. $\downarrow$ = lower is better, $\uparrow$ = higher is better.}
\label{tab:sota_comparison}
\resizebox{0.85\textwidth}{!}{
\begin{tabular}{l|cc|cc|cc|c|cc}
\toprule
\multirow{2}{*}{Method} & \multicolumn{2}{c|}{Accuracy $\downarrow$} & \multicolumn{2}{c|}{Completeness $\downarrow$} & \multicolumn{2}{c|}{Distance $\downarrow$} & Pose $\downarrow$ & \multicolumn{2}{c}{Segmentation $\uparrow$} \\
 & Mean & Median & Mean & Median & Mean & Median & ATE & JM & FM \\
\midrule
MonST3R & 0.0378 & 0.0256 & 0.1034 & 0.0698 & 0.0721 & 0.0489 & 0.0198 & 0.0189 & 0.1124 \\
DAS3R  & 0.0362 & 0.0241 & 0.0995 & 0.0672 & 0.0698 & 0.0465 & 0.0189 & 0.0198 & 0.1187 \\
CUT3R  & 0.0355 & 0.0237 & 0.0978 & 0.0658 & 0.0678 & 0.0452 & 0.0185 & 0.0215 & 0.1232 \\
Easi3R  & 0.0425 & 0.0298 & 0.1156 & 0.0812 & 0.0845 & 0.0598 & 0.0215 & 0.0156 & 0.0987 \\
VGGT4D  & \underline{0.0350} & \underline{0.0233} & \underline{0.0967} & \underline{0.0641} & \underline{0.0659} & \underline{0.0437} & \underline{0.0182} & \underline{0.0207} & \underline{0.1249} \\
\midrule
\textbf{Ours} & \textbf{0.0303} & \textbf{0.0210} & \textbf{0.0864} & \textbf{0.0550} & \textbf{0.0583} & \textbf{0.0380} & \textbf{0.0181} & \textbf{0.0226} & \textbf{0.1380} \\
\bottomrule
\end{tabular}
}
\end{table*}

\begin{table}[h]
\centering
\caption{Comparison on Comparison with state-of-the-art methods on DAVIS-2016 dataset. Best results in bold, second best underlined. $\downarrow$ = lower is better, $\uparrow$ = higher is better.}
\label{tab:davis_comparison}
\begin{tabular}{l|cccc}
\toprule
\multirow{2}{*}{Method} & \multicolumn{4}{c}{DAVIS-2016} \\ \cmidrule{2-5} 
 & JM$\uparrow$ & JR$\uparrow$ & FM$\uparrow$ & FR$\uparrow$ \\ \midrule
Easi3R$_{\text{dust3r}}$ & 50.10 & 55.77 & 43.40 & 37.25 \\
Easi3R$_{\text{monst3r}}$ & 54.93 & 68.00 & 45.29 & 47.30 \\
MonST3R & 40.42 & 40.39 & 49.54 & 52.12 \\
DAS3R & 41.13 & 38.67 & 44.50 & 36.94 \\ 
VGGT4D & \underline{60.19} & \textbf{78.39} & \underline{54.81} & \underline{67.49} \\ \midrule
\textbf{Ours} & \textbf{61.60} & \underline{76.70} & \textbf{55.47} & \textbf{67.52} \\
\bottomrule
\end{tabular}
\end{table}

\subsection{Comparison with State-of-the-Art}

We evaluate our framework against various state-of-the-art methods, including training-based approaches like MonST3R~\cite{monst3r}, DAS3R~\cite{das3r}, and CUT3R~\cite{cut3r}, as well as our primary baseline VGGT4D~\cite{vggt4d}. As shown in Table~\ref{tab:sota_comparison}, our method achieves superior performance on the DyCheck \cite{dycheck} dataset across all evaluation dimensions. Notably, we achieve a significant reduction in reconstruction errors, with Accuracy Mean improving from 0.0350 to 0.0303 and Completeness Median from 0.0641 to 0.0550. Our approach also demonstrates more robust pose estimation (ATE: 0.0181) and superior dynamic segmentation (FM: 0.1380) compared to methods explicitly trained on dynamic sequences.

On the DAVIS-2016 \cite{davis2016} benchmark (Table~\ref{tab:davis_comparison}), our framework consistently outperforms existing methods in segmentation quality. We achieve the best scores in Jaccard Mean (61.60) and Boundary F-measure (55.47). These results indicate that our hierarchical uncertainty modeling effectively isolates dynamic motion cues, enabling more precise object-background separation than previous transformer-based architectures. The fact that we consistently surpass training-heavy baselines like MonST3R and DAS3R validates that principled uncertainty-aware weighting can effectively compensate for the absence of task-specific training data, establishing a new state-of-the-art for pose-free 4D reconstruction.

\begin{figure*}[ht]
  \centering
  \includegraphics[width=1\linewidth]{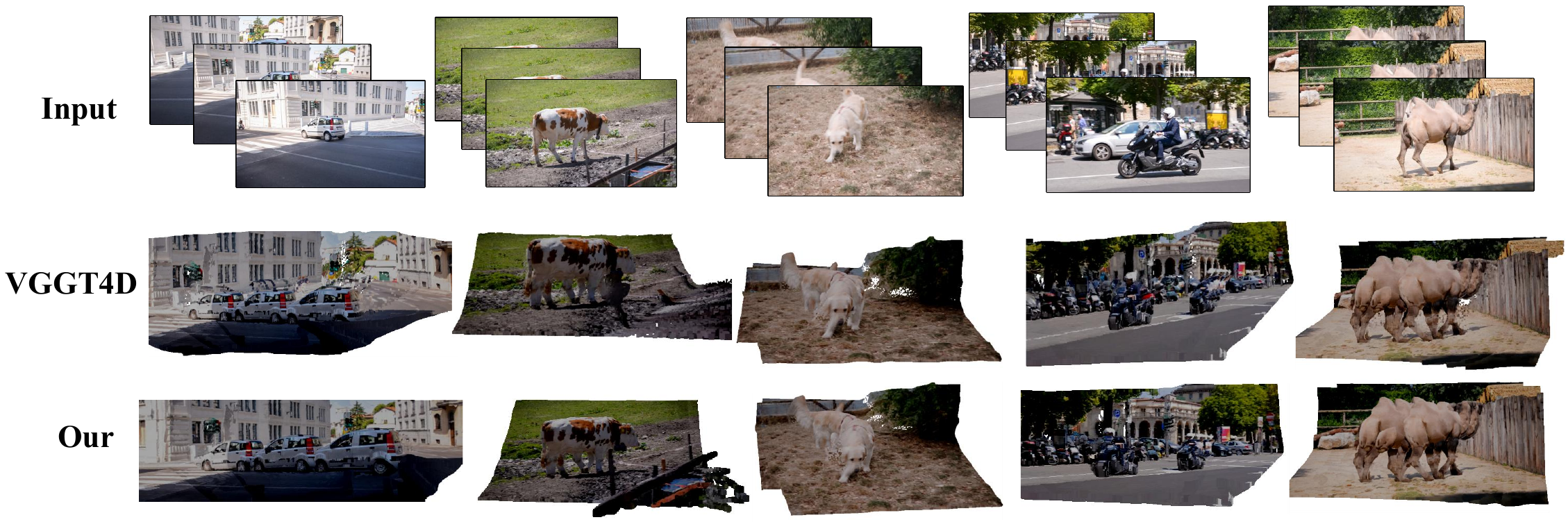} 
  \caption{Qualitative comparisons of 4D reconstruction.}
  \label{fig:vis1}
\end{figure*}

\begin{figure*}[ht]
  \centering
  \includegraphics[width=1\linewidth]{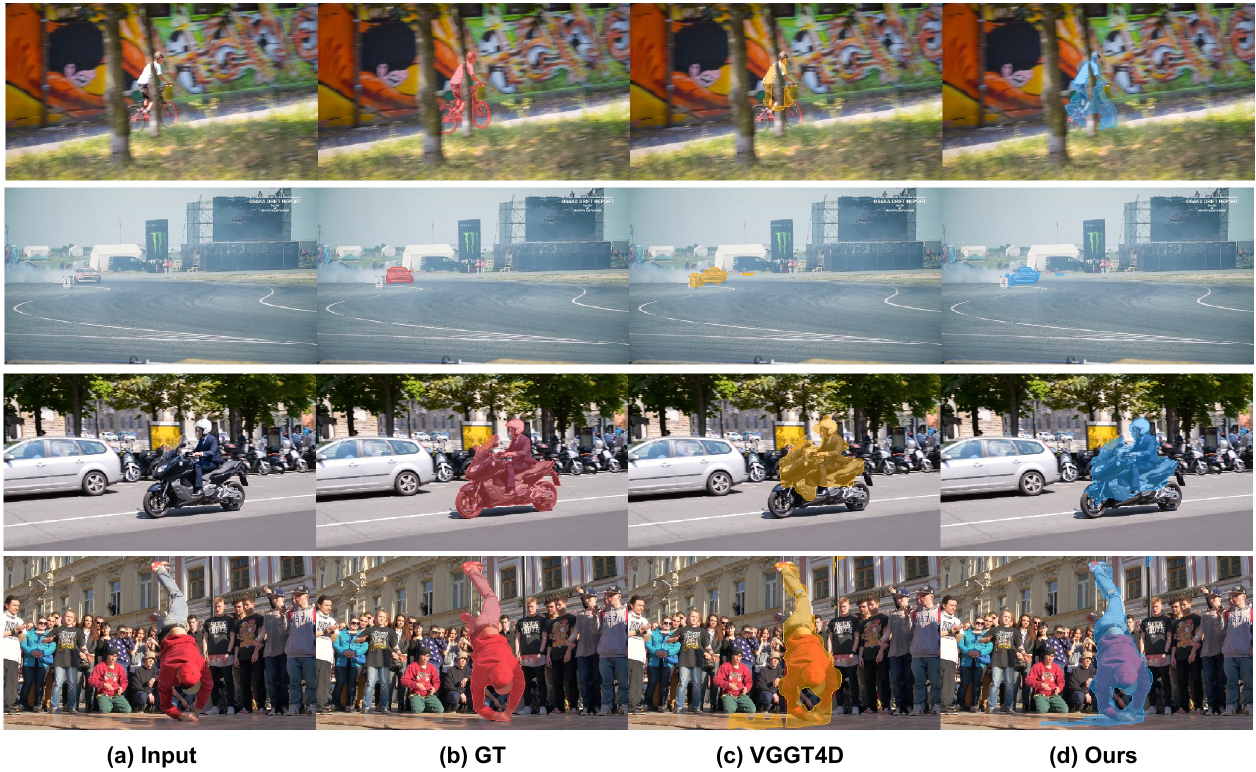} 
  \caption{Qualitative results of dynamic object segmentation.}
  \label{fig:vis2}
\end{figure*}

\begin{table*}[t]
\centering
\caption{Ablation study in DyCheck dataset. We progressively add each mechanism to the baseline and measure the cumulative improvement across corresponding benchmarks. $\downarrow$ = lower is better, $\uparrow$ = higher is better. }
\label{tab:ablation}
\resizebox{\textwidth}{!}{
\begin{tabular}{l|cc|cc|cc|c|cc}
\toprule
\multirow{2}{*}{Configuration} & \multicolumn{2}{c|}{Accuracy $\downarrow$} & \multicolumn{2}{c|}{Completeness $\downarrow$} & \multicolumn{2}{c|}{Distance $\downarrow$} & Pose $\downarrow$ & \multicolumn{2}{c}{Segmentation $\uparrow$} \\
 & Mean & Median & Mean & Median & Mean & Median & ATE & JM & FM \\
\midrule
Baseline & 0.0350 & 0.0233 & 0.0967 & 0.0641 & 0.0659 & 0.0437 & 0.0182 & 0.0207 & 0.1249 \\
+ Uncertainty-Aware Cross-View Consistency & 0.0308 & 0.0215 & 0.0866 & 0.0556 & 0.0587 & 0.0386 & 0.0182 & 0.0208 & 0.1255 \\
+ Local-Consistency Driven Geometry Purification & 0.0307 & 0.0215 & 0.0869 & 0.0560 & 0.0588 & 0.0388 & 0.0181 & 0.0209 & 0.1257 \\
+ Entropy-Guided Subspace Projection & \textbf{0.0303} & \textbf{0.0210} & \textbf{0.0864} & \textbf{0.0550} & \textbf{0.0583} & \textbf{0.0380} & \textbf{0.0181} & \textbf{0.0226} & \textbf{0.1380} \\
\midrule
\textit{Total Improvement} & \textit{13.43\%} & \textit{9.87\%} & \textit{10.65\%} & \textit{14.20\%} & \textit{11.53\%} & \textit{13.04\%} & \textit{0.55\%} & \textit{9.18\%} & \textit{10.49\%} \\
\bottomrule
\end{tabular}
}
\end{table*}

\subsection{Qualitative Analysis}

The qualitative comparisons in Figure \ref{fig:vis1} highlight the robustness of our framework in complex 4D scenarios. As shown in the comparative visualizations, the baseline VGGT4D exhibits significant geometric artifacts, where dynamic objects often appear "ghosted" or smeared across the static environment due to incomplete temporal decoupling. In contrast, our method produces visibly cleaner static backgrounds and maintains the structural integrity of dynamic objects, such as the moving vehicles and animals shown in the various sequences. By explicitly modeling hierarchical uncertainty, our approach effectively suppresses structural outliers and "floater" artifacts, resulting in more coherent 4D reconstructions with sharp boundaries and well-separated motion components.

The qualitative results in Figure \ref{fig:vis2} demonstrate the superior segmentation performance of our method. Compared to the baseline VGGT4D, which often produces over-segmented masks that bleed into static background regions or fails to capture complete moving objects, our approach yields significantly more precise masks. By leveraging hierarchical uncertainty modeling, our method effectively isolates dynamic components with sharper boundaries and higher structural integrity across various challenging scenarios, such as fast-moving vehicles and complex human articulations.

\subsection{Ablation Studies}

Table \ref{tab:ablation} presents the ablation results validating each
mechanism’s contribution.

\textbf{Uncertainty-Aware Cross-View Consistency:} The introduction of uncertainty-aware projection weighting yields
the most dramatic improvement in geometric metrics:
12.00\% reduction in Accuracy Mean and 13.26\% reduction in Completeness Median. This empirically validates
our heteroscedastic formulation: by weighting observations
according to their confidence, we suppress the influence of
unreliable depth predictions. The improvement is most pronounced in occluded and textureless regions, where the traditional uniform-weight approach introduces systematic errors.

\textbf{Local-Consistency Driven Geometry Purification:}
The geometry purification mechanism provides a stable regularization effect, with a 0.29\% additional reduction in Accuracy Mean and a 0.55\% improvement in Pose ATE. While
the quantitative gains appear modest in isolation, this mechanism plays a critical structural role: it prevents outlier
points from corrupting the spatial clustering and subsequent
projection analysis.

\textbf{Entropy-Guided Subspace Projection:} The entropyguided subspace projection delivers the strongest improvement in segmentation quality: 9.18\% in JM and 10.49\% in FM. This validates our formulation: by projecting the multi-head attention distribution onto the low-entropy subspace,
we effectively amplify the dynamic motion cues while suppressing noisy heads. The substantial improvement in F-Measure indicates sharper mask boundaries, which directly translates to reduced “floater” artifacts in the 4D reconstruction.

\section{Conclusion and Discussion}

We have presented a framework
for robust 4D scene reconstruction that can be rapidly implemented to existing Visual Geometry Transformer. Our key insight is that
dynamic-static decoupling requires principled uncertainty
modeling across multiple abstraction levels. Building on
this, we proposed a unified hierarchical framework that addresses the problem across three levels: feature representation (via entropy-weighted attention aggregation), geometric structure (via radius-based point cloud purification),
and projection constraints (via confidence-weighted heteroscedastic loss).  Extensive experiments across multiple dynamic
benchmarks validate our method with state-of-the-art performance.

\clearpage

\bibliographystyle{unsrt}
\bibliography{main}

\clearpage

\end{document}